\documentclass{article}

\usepackage[utf8]{inputenc}
\usepackage[english]{babel}
\usepackage{amsmath, amssymb}
\usepackage{booktabs}
\usepackage{graphicx}
\usepackage{hyperref}
\usepackage[ruled,vlined]{algorithm2e}
\usepackage{xcolor}
\usepackage{subcaption}
\usepackage{multirow}
\usepackage{float}  
\usepackage[a4paper, margin=2.5cm]{geometry}
\usepackage{natbib}

\title{PEFT of SLM for Telecommunications Customer Support: A Comparative Study of LoRA Configurations with Energy Consumption Analysis}

\author{
    Lucas Tamic, Ilan Jaffeux-Cheniout, Xavier Marjou \\
  Orange \\
  \texttt{\{lucas.tamic, ilan.jaffeux{-}{-}cheniout, xavier.marjou\}@orange.com}
}
\date{}

\begin{document}

\maketitle

\begin{abstract}
While LLMs have demonstrated remarkable capabilities in natural language understanding and generation, their systematic evaluation and adaptation to domain-specific constraints in telecommunications customer support remain relatively limited in the academic literature. Moreover, issues related to data sovereignty, regulatory compliance, and the protection of sensitive customer and network information further complicate the deployment of externally hosted foundation models in this sector. We present a systematic study of parameter-efficient fine-tuning (PEFT) using Low-Rank Adaptation (LoRA) on Qwen2.5-3B to develop a domain-specific conversational assistant. Our methodology introduces a combinatorial approach to synthetic dataset generation, leveraging a glossary of 52 industry-specific terms to produce approximately 30,000 training examples covering 1,560 distinct problem scenarios via a generative pipeline using Gemini 2.0 Flash. We conduct a comprehensive empirical evaluation of 16 distinct LoRA configurations, systematically varying hyperparameters and target modules. Critically, we extend traditional performance metrics to include both energy consumption analysis and qualitative evaluation using LLM-as-a-judge methodology with GPT-5.2 and Claude 4.5 Sonnet. Our findings reveal a striking divergence: fine-tuned configurations of the model with the lowest validation loss do not achieve the highest qualitative rankings. The best quantitative performer (validation loss 0.5024) ranks 6th-7th qualitatively, while the worst quantitative performer (loss 0.6807) ranks 1st by both human-aligned judges. This work contributes: (1) a novel combinatorial methodology for synthetic data construction, (2) an analysis of the relevance of target module selection for the injection of LoRA parameters, (3) demonstration that validation loss alone is insufficient for selecting among different fine-tuning configurations of the model in conversational AI, and (4) systematic energy-performance trade-off analysis for sustainable LLM deployment.
\\
\end{abstract}

\bigskip

\section{Introduction}

The deployment of language models in specialized industrial domains requires addressing challenges that go beyond general-purpose natural language processing. Telecommunications customer support is a particularly demanding application, requiring precise technical knowledge, domain-specific terminology, and structured troubleshooting procedures. Fine-tuning smaller language models (SLMs) provides a practical and efficient solution to these requirements. Through parameter-efficient adaptation, such models can be rapidly specialized using only a few thousand domain-specific examples. This approach enables operators to deploy fully customized conversational assistants hosted on their own infrastructure, ensuring data confidentiality, reducing operational latency, and allowing specialization across a wide range of technical domains.

Parameter-efficient fine-tuning (PEFT) methods, particularly Low-Rank Adaptation (LoRA)~\cite{hu2021lora}, offer a practical path forward. LoRA introduces trainable low-rank matrices into specific model layers while freezing pre-trained parameters, enabling domain adaptation with minimal computational overhead. However, several critical research questions remain: How should synthetic training data be constructed when domain-specific conversations are scarce? Which LoRA configurations optimize performance for conversational tasks? How do quantitative metrics (loss, perplexity) relate to qualitative response quality? What are the energy implications of different configurations?
\\
\\
\\

This work addresses these questions through systematic experimentation on Qwen2.5-3B for telecommunications support. Specifically, we evaluate 16 distinct LoRA-based fine-tuning configurations of the same underlying model, keeping architecture and weight count constant while varying adaptation strategies. We make four key contributions:

\begin{itemize}
    \item \textbf{Combinatorial synthetic dataset generation}: A structured methodology factorizing 52 technical terms across problem causes and usage contexts, systematically generating 1,560 distinct problem scenarios via LLM-based expansion.

    \item \textbf{Comprehensive LoRA configuration study}: Empirical evaluation of 16 configurations varying learning rates, batch sizes, ranks (16, 32), and target module sets (2, 4, 7 modules), showing that increasing the number of LoRA parameters through broader target module coverage has a more pronounced impact on reducing loss and perplexity than increasing the number of parameters by raising the LoRA rank.

    \item \textbf{Quantitative-qualitative performance divergence}: LLM-as-a-judge evaluation with GPT-5.2 and Claude 4.5 Sonnet shows that fine-tuning configurations with lower validation loss do not necessarily achieve higher qualitative rankings, whereas configurations with higher validation loss consistently rank better in qualitative assessments.

    \item \textbf{Energy-aware evaluation}: Comprehensive power consumption measurements (284-1371 Wh) show configuration choices impact energy by 5$\times$, with performance-efficiency sweet spots emerging from careful hyperparameter selection.
\end{itemize}

The remainder of this paper is organized as follows: Section~\ref{sec:related} reviews related work. Section~\ref{sec:methodology} describes dataset construction, model architecture, and LoRA configurations. Section~\ref{sec:experiments} details experimental setup. Section~\ref{sec:results} presents quantitative and qualitative results with divergence analysis. Section~\ref{sec:conclusion} concludes with implications and future directions.

\section{State of the Art}
\label{sec:related}

\subsection{Parameter-Efficient Fine-Tuning}

Parameter-efficient fine-tuning methods address the computational challenges of adapting pre-trained models. LoRA~\cite{hu2021lora} decomposes weight updates into low-rank matrices, reducing trainable parameters while maintaining competitive performance. Variants include AdaLoRA~\cite{zhang2023adalora} with adaptive rank allocation, and QLoRA~\cite{dettmers2023qlora} combining LoRA with 4-bit quantization. Alternative approaches include prefix tuning~\cite{li2021prefix} and prompt tuning~\cite{lester2021prompt}. Our work focuses on standard LoRA, systematically studying the impact of hyperparameters on conversational AI performance as well as on the energy consumption during training.

\subsection{Synthetic Data Generation}

Data scarcity in specialized domains has driven synthetic data generation research. Self-Instruct~\cite{wang2023selfinstruct} and Alpaca~\cite{taori2023alpaca} demonstrated LLMs can bootstrap instruction-following via self-generation. Our approach differs by introducing explicit combinatorial structure—factorizing domain knowledge into orthogonal dimensions enables controlled problem space exploration while maintaining diversity.

\subsection{Energy Efficiency and Evaluation}

Recent work documents substantial carbon footprints of training large models~\cite{strubell2019energy,patterson2021carbon}, advocating for Green AI practices~\cite{schwartz2020green}. However, energy analysis remains uncommon in PEFT research. Simultaneously, emerging work on LLM-as-a-judge evaluation~\cite{baysan2025llmjudge} shows that model-based evaluation can approximate human judgments for response quality assessment. Our work uniquely combines energy measurement and qualitative evaluation to provide a holistic assessment of the model’s fine-tuning configurations, beyond traditional loss and perplexity metrics.
\\
\\
\\
\\
\\
\\

\section{Methodology}
\label{sec:methodology}

\subsection{Synthetic Dataset Generation}

Public telecommunications support datasets are scarce and operator-specific logs raise privacy concerns. We address this through structured synthetic generation combining domain knowledge engineering with LLM-based expansion.

\subsubsection{Domain Knowledge Factorization}

We begin with a glossary of 52 technical terms from a major European telecommunications operator (Orange) spanning network technologies, services, business models, technical components, and infrastructure. Rather than unstructured generation, we factorize the problem space into orthogonal dimensions:

\begin{itemize}
    \item \textbf{Technical terms} ($T$): 52 glossary concepts
    \item \textbf{Problem causes} ($C$): Approximately 10 common failure modes per term (configuration errors, interference, congestion, equipment faults)
    \item \textbf{Usage contexts} ($L$): Approximately 3 location/situation contexts per term (home, office, public spaces, specific conditions)
\end{itemize}

This factorization yields a combinatorial problem space:
\begin{equation}
|T| \times |C| \times |L| \approx 52 \times 10 \times 3 \approx 1{,}560 \text{ distinct problem scenarios}
\end{equation}

By decomposing the problem into three components, we promote diversity in the generated examples, preventing the model from repeatedly producing similar scenarios.

\subsubsection{LLM-Based Generation Pipeline}

Given the combinatorial space, we employ Gemini 2.0 Flash~\cite{google2024gemini} to generate natural language examples. The pipeline operates in two phases:

\textbf{User Query Generation:} For each sampled combination $(t, c, l)$, we generate realistic customer queries incorporating the technical term $t$, implying problem cause $c$, and specifying context $l$ with linguistic diversity avoiding templatic patterns.

\textbf{Expert Response Generation:} For each query, we produce responses following a two-stage resolution strategy: (1) self-service troubleshooting with 3-5 actionable steps the user can perform independently, and (2) escalation recommendation if self-service fails, explicitly suggesting human support contact.
\\
\\
\\
\\
\\
\\
\\
\\
\\
\\
\\
\\
\\
\\
\\
\\
\\
\subsubsection{Dataset Characteristics}

We generated 30,000 conversational examples in JSONL format, each containing: (1) \textbf{Instruction}: Fixed system prompt defining the assistant role, (2) \textbf{Input}: User query, (3) \textbf{Output}: Expert response. Example:

\begin{verbatim}
{
    "Instruction": "As a telecom expert, answer network-related 
    questions with actions the user can perform themselves.", 
    
    "Input": "My 4G keeps disconnecting and reconnecting in 
    this shopping mall. I have to constantly reset my network. What can I do?", 
    
    "Output": "Frequent network resets in a crowded area like 
    a shopping mall can be due to several factors. 
    Here are some things you can try:
    
    1. Move to a less crowded area within the mall: 
    Network congestion is common in high-traffic locations.
    
    2. Restart your phone: This can sometimes resolve temporary 
    software glitches affecting network connectivity.
    
    3. Toggle Airplane Mode: Turn Airplane Mode on for about 15 seconds,
    then turn it off. This forces your phone to reconnect to the network.
    
    4. Check your SIM card: Ensure your SIM card is properly inserted 
    and clean. 
    A loose or dirty SIM card can cause intermittent connection issues.
    
    5. Temporarily switch to 3G/2G (if possible): While slower, 
    these networks might be less congested and more stable in that 
    specific location. Check your phone's settings for this option.
    
    If these steps do not resolve the problem, contact your mobile 
    operator as the issue could be due to network infrastructure."}

\end{verbatim}

The dataset exhibits technical coverage across all 52 terms, linguistic diversity in phrasing and sophistication, consistent resolution structure, and JSONL formatting for straightforward integration into training pipelines.
\\
\\
\subsection{Base Model and LoRA Configuration}

We employ Qwen2.5-3B~\cite{yang2024qwen25} as our base model. The 3B parameter scale balances representational capacity for domain-specific fine-tuning with computational tractability and deployment feasibility. Qwen2.5-3B uses a standard decoder-only transformer with 36 layers, 2,048 hidden dimensions, 16 attention heads, Grouped-Query Attention (GQA) with 2 key-value heads, and 151,936 vocabulary tokens.

For LoRA~\cite{hu2021lora}, we introduce trainable low-rank matrices: $W = W_0 + BA$ where $B \in \mathbb{R}^{d \times r}$, $A \in \mathbb{R}^{r \times k}$, and $r \ll \min(d, k)$ is the rank. Trainable parameters: $r \times (d + k)$ versus $d \times k$ for full fine-tuning. LoRA can target different transformer modules:
\begin{itemize}
    \item \textbf{Attention projections}: Query ($W_q$), Key ($W_k$), Value ($W_v$), Output ($W_o$)
    \item \textbf{Feed-forward layers}: Gate ($W_{\text{gate}}$), Up ($W_{\text{up}}$), Down ($W_{\text{down}}$)
    \\
    
\end{itemize}

\subsection{Configuration Grid}

We systematically evaluate 16 configurations varying:
\begin{itemize}
    \item \textbf{Learning rate}: $\{5\times10^{-5}, 1\times10^{-4}, 2\times10^{-4}\}$
    \item \textbf{Effective batch size}: $\{4, 8, 16, 32\}$ via per-device batch size and gradient accumulation
    \item \textbf{LoRA rank ($r$)}: $\{16, 32\}$ with $\alpha = 2r$
    \item \textbf{Target modules}:
    \begin{itemize}
        \item Minimal (2 modules): $\{q\_proj, v\_proj\}$
        \item Moderate (4 modules): $\{q\_proj, v\_proj, k\_proj, o\_proj\}$
        \item Extensive (7 modules): $\{q\_proj, v\_proj, k\_proj, o\_proj, gate\_proj, up\_proj, down\_proj\}$
    \end{itemize}
\end{itemize}

\begin{table}[H]
\centering
\caption{Hyperparameter Settings for 16 Fine-Tuning Variants of the Model.}
\label{tab:lora_configs}
\resizebox{\textwidth}{!}{%
\begin{tabular}{@{}ccccccc@{}}
\toprule
\textbf{Configuration} & \textbf{Batch Size} & \textbf{Grad. Accum.} & \textbf{Learning Rate} & \textbf{r} & \textbf{Alpha} & \textbf{Target Modules} \\
\midrule
1  & 1 & 8  & 1e-4 & 16 & 32 & q, v \\
2  & 1 & 4  & 1e-4 & 16 & 32 & q, v \\
3  & 2 & 4  & 1e-4 & 16 & 32 & q, v \\
4  & 2 & 4  & 1e-4 & 16 & 32 & q, v, k, o \\
5  & 1 & 8  & 2e-4 & 16 & 32 & q, v, k, o \\
6  & 1 & 8  & 5e-5 & 16 & 32 & q, v, k, o \\
7  & 1 & 8  & 1e-4 & 16 & 32 & q, v, k, o \\
8  & 1 & 8  & 1e-4 & 16 & 32 & q, v, k, o, gate, up, down \\
9  & 1 & 8  & 5e-5 & 16 & 32 & q, v, k, o, gate, up, down \\
10 & 1 & 8  & 2e-4 & 16 & 32 & q, v, k, o, gate, up, down \\
11 & 1 & 8  & 1e-4 & 32 & 64 & q, v, k, o \\
12 & 1 & 8  & 1e-4 & 32 & 64 & q, v, k, o, gate, up, down \\
13 & 1 & 8  & 2e-4 & 32 & 64 & q, v, k, o, gate, up, down \\
14 & 1 & 16 & 1e-4 & 16 & 32 & q, v \\
15 & 1 & 32 & 1e-4 & 16 & 32 & q, v \\
16 & 1 & 8  & 5e-5 & 32 & 64 & q, v, k, o, gate, up, down \\
\bottomrule
\end{tabular}%
}
\end{table}

\subsection{Training Protocol}

\textbf{Data preparation:} The 30,000 examples are tokenized using Qwen's tokenizer with custom special tokens (\texttt{<|user|>}, \texttt{<|assistant|>}). Each example is formatted as:
\begin{verbatim}
<|user|>{Instruction} {Input}<|enduser|>
<|assistant|>{Output}<|endassistant|>
\end{verbatim}

We apply completion-only loss masking, computing loss only on assistant response tokens to focus learning on response generation rather than input prediction.

\textbf{Data split:} 95\% training (28,500 examples), 5\% validation (1,500 examples) with fixed random seed (42).

\textbf{Optimization:} AdamW~\cite{loshchilov2019adamw} with weight decay 0.01, maximum sequence length 2,048 tokens, maximum 10 epochs, bfloat16 mixed precision.

\textbf{Early stopping:} Patience of 3 evaluation steps monitoring validation loss, retaining the checkpoint with lowest validation loss.

\textbf{Evaluation metrics:} Cross-entropy loss and perplexity (exponential of loss) computed on validation set.
\\
\\
\\
\\
\section{Experiments}
\label{sec:experiments}

\subsection{Experimental Setup}

All experiments are conducted on a single NVIDIA GPU RTX 4090. Training employs PyTorch with Transformers~\cite{wolf2020transformers}, PEFT~\cite{mangrulkar2022peft}, and TRL~\cite{vonwerra2022trl} libraries. The fine-tuned model is periodically evaluated every 100 training steps, and the version that attains the best performance during training is retained. Early stopping halts training if the validation loss fails to improve for 3 consecutive evaluations.

\subsection{Energy Measurement}

We measure GPU power consumption using NVIDIA's \texttt{nvidia-smi}, polling power draw at 1-second intervals throughout training. For each configuration, we record: training duration (wall-clock time), average GPU power (watts), and total energy consumption (watt-hours). Total energy: $E_{\text{total}} = \frac{1}{3600} \sum_{i=1}^{N} P_i \Delta t_i$ where $P_i$ is power reading $i$ and $\Delta t_i \approx 1$ second. This captures GPU energy but excludes CPU, memory, storage, and cooling infrastructure.

\subsection{Qualitative Evaluation Methodology}

To complement quantitative metrics, we conduct qualitative evaluation using the LLM as a judge paradigm~\cite{zheng2023judging}. We generate responses from all 16 fine-tuned models plus the base model for 9 diverse telecommunications support prompts covering network connectivity (MMS, SMS, signal), service-specific problems (voicemail, eSIM, VOD), performance degradation (latency, call dropping), and infrastructure challenges (small cells, VDSL sync).

Two state-of-the-art LLMs independently rank all 17 model variants:
\begin{itemize}
    \item \textbf{GPT-5.2} (OpenAI)
    \item \textbf{Claude 4.5 Sonnet} (Anthropic)
\end{itemize}

Evaluation criteria (implicit in judge prompts): technical accuracy and relevance, clarity and structure of troubleshooting steps, appropriateness of escalation recommendations, and natural language quality and tone. Each judge provides a complete ranking from 1 (best) to 17 (worst) based on overall response quality across all 9 prompts.

\section{Results}
\label{sec:results}

\subsection{Quantitative Performance}

\begin{table}[H]
\centering
\caption{Complete results for all 16 LoRA configurations including quantitative metrics, LoRA parameters, energy consumption, and qualitative rankings. Rank indicates position by validation loss (1=best). Duration format: hours:minutes:seconds.}
\label{tab:complete_results}
\begin{tabular}{@{}cccccc@{}}
\toprule
\textbf{Rank} & \textbf{Configuration} & \textbf{Validation Loss} & \textbf{Perplexity} & \textbf{r} & \textbf{Modules}  \\
\midrule
1  & configuration 8  & 0.5024 & 1.653 & 16 & 7 (all)  \\
2  & configuration 16 & 0.5146 & 1.673 & 32 & 7 (all)  \\
3  & configuration 13 & 0.5516 & 1.736 & 32 & 7 (all)  \\
4  & configuration 9  & 0.5559 & 1.744 & 16 & 7 (all)  \\
5  & configuration 12 & 0.5613 & 1.753 & 32 & 7 (all)  \\
6  & configuration 10 & 0.5630 & 1.756 & 16 & 7 (all)  \\
7  & configuration 5  & 0.5714 & 1.771 & 16 & 4 (q,k,v,o)  \\
8  & configuration 11 & 0.5800 & 1.786 & 32 & 4 (q,k,v,o)  \\
\midrule
9  & configuration 6  & 0.6228 & 1.864 & 16 & 4 (q,k,v,o)  \\
10 & configuration 7  & 0.6235 & 1.866 & 16 & 4 (q,k,v,o)  \\
11 & configuration 4  & 0.6239 & 1.866 & 16 & 4 (q,k,v,o)  \\
12 & configuration 3  & 0.6526 & 1.921 & 16 & 2 (q,v)  \\
13 & configuration 1  & 0.6532 & 1.922 & 16 & 2 (q,v)  \\
14 & configuration 15 & 0.6694 & 1.953 & 16 & 2 (q,v)  \\
15 & configuration 14 & 0.6700 & 1.954 & 16 & 2 (q,v)  \\
16 & configuration 2  & 0.6807 & 1.975 & 16 & 2 (q,v)  \\
\bottomrule
\end{tabular}
\end{table}

\subsubsection{Target Module Selection Dominates Performance}

The most striking finding is that target module selection strongly influences validation performance. All top-6 configurations (loss 0.50–0.57) apply LoRA to all 7 modules (attention plus feed-forward), whereas configurations targeting only 2 attention modules consistently show higher losses (0.65–0.68). Configurations with 4 modules achieve intermediate performance. These results indicate that, in order to achieve lower validation loss, it is particularly important to include the feed-forward layers in the adaptation process. Feed-forward layers, hypothesized to store factual knowledge~\cite{geva2021transformer}, seem to play a key role in integrating domain-specific terminology and troubleshooting procedures, although this does not guarantee overall fine-tuning quality.

Targeting the feed-forward (FFN) layers in LoRA presents a compelling opportunity for domain-specific fine-tuning. FFN layers are highly expressive and capable of storing and transforming factual knowledge, making them particularly effective for integrating domain-specific terminology and complex reasoning patterns. However, this expressiveness also introduces a risk: when the training dataset is relatively small or consists of short, simple examples, the model can adapt too strongly to these examples, leading to overfitting. To fully leverage the potential of FFN adaptation, a sufficiently complex and diverse dataset is desirable, ideally containing relatively long sequences and a variety of scenarios that encourage the model to generalize rather than simply memorize.

\subsubsection{LoRA Rank: r=16 Outperforms r=32}

Contrary to intuitive expectations, increasing the LoRA rank does not necessarily improve validation performance. One might assume that a higher rank, by providing more LoRA parameters, would increase the model's capacity to learn. However, our results show otherwise:
\begin{itemize}
    \item configuration 8 (r=16, 7 modules): loss 0.5024 (rank 1)
    \item configuration 12 (r=32, 7 modules): loss 0.5613 (rank 5)
\end{itemize}

These results suggest that r=16 provides sufficient capacity for this domain adaptation task, whereas r=32 offers diminishing or even negative returns. Several factors may explain this phenomenon: (1) r=16 adequately captures the effective representation space needed for conversational telecommunications support, (2) r=32 may introduce a slight overfitting risk due to excessive degrees of freedom, (3) lower ranks maintain a higher signal-to-noise ratio by focusing on the primary adaptation directions within 16 dimensions, and (4) faster convergence with r=16 allows better generalization before early stopping is triggered.

\subsubsection{Impact of Target Module vs LoRA Rank on Loss Reduction}

Our results indicate that increasing the number of LoRA parameters by expanding target module coverage is more effective for reducing validation loss than increasing the number of parameters by raising the LoRA rank. While both strategies increase the overall adaptation capacity, they affect the model in fundamentally different ways.

Expanding the set of target modules distributes LoRA adaptations across a broader range of functional components of the transformer, including both attention mechanisms and feed-forward layers. This allows the model to adjust multiple stages of representation processing, leading to more effective error correction and a more uniform reduction of loss across layers. However, increasing the LoRA rank means adding more parameters to the same modules of the model. This primarily strengthens the model’s ability to represent certain types of patterns or relationships along directions that it already handles well at a lower rank. In other words, these additional parameters do not necessarily provide new learning capacity but rather amplify what the model has already learned.

From a loss optimization perspective, broader module coverage enables the model to correct mismatches at multiple abstraction levels, whereas higher-rank adaptations risk introducing redundant or noisy parameter updates. 

\subsection{Qualitative Evaluation}
\label{sec:qualitative}

\begin{table}[H]
\centering
\caption{Qualitative rankings assigned by GPT-5.2 and Claude 4.5 Sonnet.}
\label{tab:qualitative_rankings}
\footnotesize 
\setlength{\tabcolsep}{10pt} 
\begin{tabular}{@{}llccc@{}}
\toprule
\textbf{Configuration} & \textbf{r} & \textbf{Modules} & \textbf{GPT-5.2 Rank} & \textbf{Claude Rank} \\
\midrule
configuration 1  & 16 & 2 (q,v)     & 6  & 5  \\
configuration 2  & 16 & 2 (q,v)     & \textbf{1}  & \textbf{1}  \\
configuration 3  & 16 & 2 (q,v)     & 2  & 3  \\
configuration 4  & 16 & 4 (q,k,v,o) & 3  & 2  \\
configuration 5  & 16 & 4 (q,k,v,o) & 5  & 8  \\
configuration 6  & 16 & 4 (q,k,v,o) & 4  & 4  \\
configuration 7  & 16 & 4 (q,k,v,o) & 17 & 7  \\
configuration 8  & 16 & 7 (all)     & 7  & 6  \\
configuration 9  & 16 & 7 (all)     & 8  & 9  \\
configuration 10 & 16 & 7 (all)     & 9  & 10 \\
configuration 11 & 32 & 4 (q,k,v,o) & 10 & 11 \\
configuration 12 & 32 & 7 (all)     & 11 & 13 \\
configuration 13 & 32 & 7 (all)     & 12 & 14 \\
configuration 14 & 16 & 2 (q,v)     & 13 & 12 \\
configuration 15 & 16 & 2 (q,v)     & 14 & 15 \\
configuration 16 & 32 & 7 (all)     & 15 & 16 \\
\midrule
Base model & -- & --      & 16 & 17 \\
\bottomrule
\end{tabular}
\end{table}
The base model was added to provide a reference point, highlighting the improvements achieved through training.
\bigskip
\bigskip
\bigskip
\bigskip
\bigskip
\bigskip
\bigskip
\bigskip
\subsubsection{Inter-Judge Agreement}

To assess the consistency between the qualitative rankings assigned by GPT-5.2 and Claude 4.5 Sonnet, we computed Spearman’s rank correlation coefficient. A significance test using Student’s t-distribution was then performed to determine whether the observed correlation is statistically meaningful.

\bigskip

\begin{table}[H]
\centering
\caption{Difference in Rank Selection Between GPT-5.2 and Claude 4.5 Sonnet.}
\begin{tabular}{|c|c|c|c|c|}
\hline
GPT 5.2 & Claude 4.5 Sonnet & $d$ & $d^2$ \\
\hline
configuration 2 & configuration 2 & 0 & 0 \\
configuration 3 & configuration 4 & -1 & 1 \\
configuration 4 & configuration 3 & 1 & 1 \\
configuration 6 & configuration 6 & 0 & 0 \\
configuration 5 & configuration 1 & 3 & 9 \\
configuration 1 & configuration 8 & -4 & 1 \\
configuration 8 & configuration 7 & 1 & 1 \\
configuration 9 & configuration 5 & 4 & 1 \\
configuration 10 & configuration 9 & 1 & 1 \\
configuration 11 & configuration 10 & 1 & 1 \\
configuration 12 & configuration 11 & 1 & 4 \\
configuration 13 & configuration 14 & -1 & 4 \\
configuration 14 & configuration 12 & 2 & 1 \\
configuration 15 & configuration 13 & 2 & 1 \\
configuration 16 & configuration 15 & 1 & 1 \\
base model & configuration 16 & -1 & 1 \\
configuration 7 & base model & -10 & 100 \\
\hline
\end{tabular}
\end{table} 
\bigskip
\bigskip

\noindent\textbf{Spearman's Rank Correlation Coefficient Calculation}
\\
\\
The difference between ranks is defined by:
\begin{equation}
d_i = A_i - B_i
\end{equation}
\\
\\
The sum of squared differences is:
\begin{equation}
\sum d_i^2 = 128
\end{equation}
\\
\\
Spearman's rank correlation coefficient is calculated as:
\begin{equation}
\rho = 1 - \frac{6\sum d_i^2}{n(n^2-1)}
\end{equation}
\\
\\
With $n = 17$:
\begin{equation}
\rho = 1 - \frac{6 \times 128}{17(17^2-1)} = 1 - \frac{768}{17 \times 288} = 0.843
\end{equation}

\bigskip
\noindent\textbf{Significance Test}
\\
\\
To assess whether this correlation is statistically significant, we use Student's t-test. The test statistic is:
\\
\begin{equation}
t = \rho \sqrt{\frac{n-2}{1-\rho^2}} = 0.843 \sqrt{\frac{17-2}{1-0.843^2}} = 0.843 \sqrt{\frac{15}{1-0.710}} = 6.07
\end{equation}

The degrees of freedom is:
\begin{equation}
df = n - 2 = 17 - 2 = 15
\end{equation}

\bigskip
\noindent\textbf{Conclusion}
\\
\\
For a significance level of $\alpha = 0.001$, corresponding to a 99.9\% confidence level, the critical value from the Student's t-distribution table with 15 degrees of freedom is $t_{critical} = 4.073$.

Since $t = 6.07 > 4.073$, we reject the null hypothesis that there is no correlation. Therefore, there is a statistically significant correlation between the two rankings at the 0.1\% significance level.

\subsection{Quantitative-Qualitative Performance Divergence}
\label{sec:divergence}

\begin{table}[H]
\centering
\caption{Validation loss and qualitative rankings for all 16 LoRA configurations.}
\label{tab:rankings}
\begin{tabular}{@{}cccc@{}}
\toprule
\textbf{Configuration} & \textbf{Val Loss} & \textbf{GPT-5.2 Rank} & \textbf{Claude Rank} \\
\midrule
configuration 1  & 0.6532 & 6  & 5  \\
configuration 2  & 0.6807 & \textbf{1}  & \textbf{1}  \\
configuration 3  & 0.6526 & 2  & 3  \\
configuration 4  & 0.6239 & 3  & 2  \\
configuration 5  & 0.5714 & 5  & 8  \\
configuration 6  & 0.6228 & 4  & 4  \\
configuration 7  & 0.6235 & 17 & 7  \\
configuration 8  & 0.5024 & 7  & 6  \\
configuration 9  & 0.5559 & 8  & 9  \\
configuration 10 & 0.5630 & 9  & 10 \\
configuration 11 & 0.5800 & 10 & 11 \\
configuration 12 & 0.5613 & 11 & 13 \\
configuration 13 & 0.5516 & 12 & 14 \\
configuration 14 & 0.6700 & 13 & 12 \\
configuration 15 & 0.6694 & 14 & 15 \\
configuration 16 & 0.5146 & 15 & 16 \\
\bottomrule
\end{tabular}
\end{table}

A striking divergence is observed between the quantitative and qualitative evaluations: fine-tuning configurations achieving the lowest validation loss do not consistently rank highest in the qualitative assessment.

\subsubsection{Key Observations}

\textbf{configuration 2: Worst loss, best qualitative.} Configuration 2 exhibits the highest validation loss (0.6807, rank 16/16) yet achieves 1st place by both GPT-5.2 and Claude 4.5 Sonnet—a divergence of -15 positions. This configuration targets only 2 modules (q\_proj, v\_proj) with r=16, suggesting the architectural constraint forcing adaptation through attention alone may develop more robust attention-based strategies that generalize better to human-aligned quality despite worse validation performance.

\textbf{configuration 16: Second-best loss, near-worst qualitative.} Configuration 16 achieves the 2nd-lowest validation loss (0.5146) but ranks 15th-16th qualitatively—a divergence of +13-14 positions. Despite targeting all 7 modules with r=32, responses appear less natural or appropriately calibrated from a human perspective.

\textbf{configuration 8: Best loss, medium qualitative.} Configuration 8 achieves the lowest validation loss (0.5024, rank 1) but ranks only 6th-7th qualitatively. While this represents strong performance overall, it demonstrates that minimizing validation loss does not guarantee optimal human-aligned quality.

\textbf{configurations 3-4: Strong qualitative despite higher loss.} Configuration 3 and configuration 4 rank 2nd-3rd qualitatively (both judges) despite loss ranks of 11th-12th, further confirming the divergence pattern.
\\
\\
\\
\\
\subsubsection{Hypotheses for Divergence}

Several factors may explain why some fine-tuning setups with intermediate validation loss yield higher qualitative scores than those with the lowest loss:
\\
\\
\textbf{Token-level over-optimization:} Although the training and validation sets contain different examples, the data share a similar structure, and some examples address the same problems expressed in  different ways. As a result, even a fine-tuning configuration with slight overfitting performs well on quantitative evaluation, having learned the expected response patterns. However, during qualitative evaluation, the overfitting becomes more apparent: responses may seem less natural, which is detected by the LLM judges.
\\
\\
The training and validation loss curves for configuration 8 show a sudden drop—or “step”—in the training loss around 3,500 steps, while the validation loss continues to decrease gradually. Because early stopping monitors the validation loss, which is still declining, it does not trigger, allowing this minor overfitting to persist. The same sharp drop in training loss is observed in all configurations using the 7-module configuration, explaining the slight overfitting detected during qualitative evaluation.

\begin{figure}[H]
    \centering
    \includegraphics[width=0.8\textwidth]{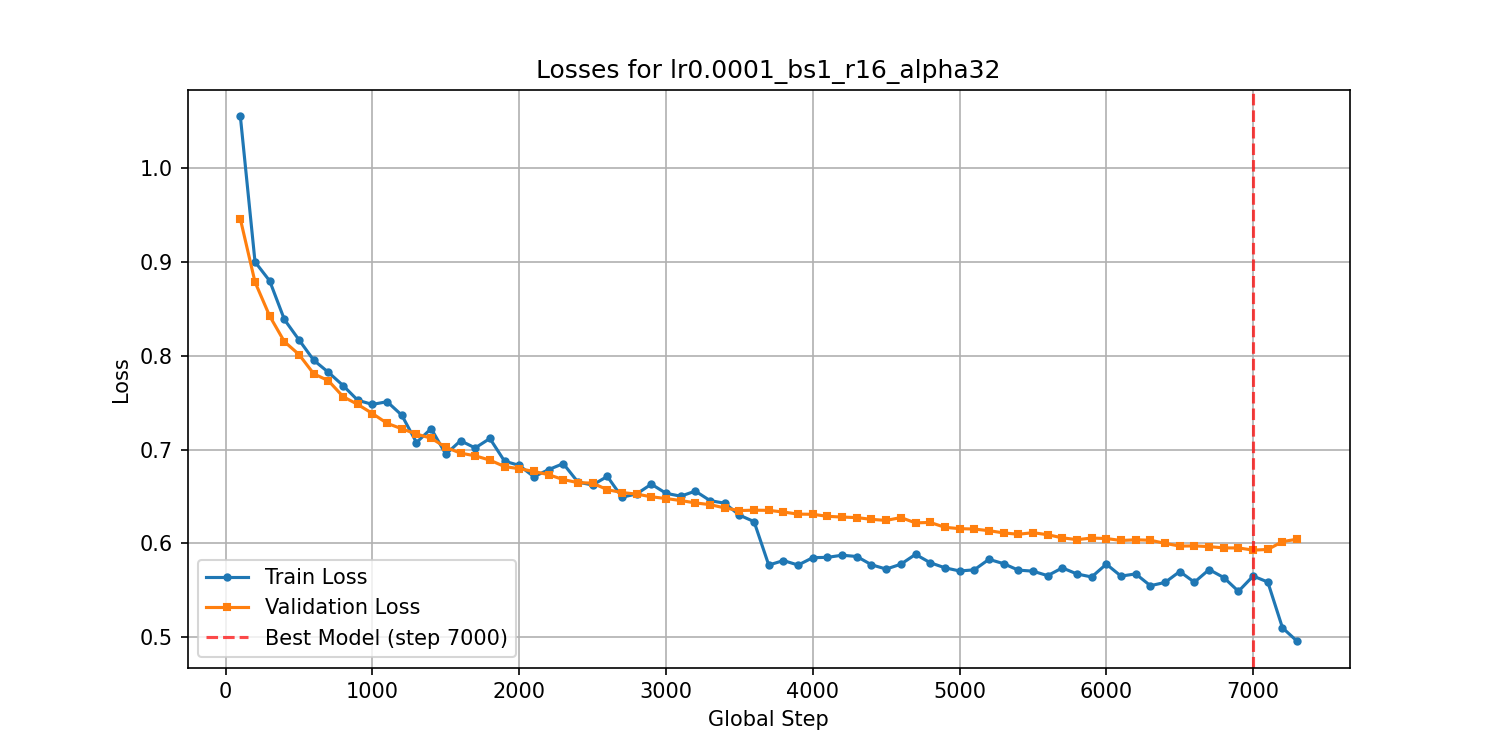}
    \caption{Training and validation loss curves for fine-tuning configuration 8, showing the evolution of the loss over training steps and the convergence behavior of this configuration.}
    \label{fig:lossucurve}
\end{figure}

Although the dataset contains a variety of examples, it does not include particularly complex cases. The examples are relatively short, averaging around 200 tokens, and they do not contain user-like errors. As a result, using all 7 attention modules during training is likely excessive. Adapting only two modules appears sufficient for this dataset. For a more complex dataset—with longer examples, greater diversity, and injected user errors—training with the feed-forward layers would likely become more beneficial, allowing the model to better capture intricate patterns.
\\
\\
\\
\\
\\
\\
\\
\\
\\
\\
\\
\\
\\
\\
\\
\\
\\

\textbf{Limitation of memorization and generation flexibility:} The configuration 2 (r = 16, targeting only two attention modules) freezes the feed-forward layers and adapts only two attention modules. This constraint limits the model’s ability to strictly memorize the structural patterns of the dataset, forcing it to reason more through attention. As a result, configuration 2 demonstrates more natural and robust responses during qualitative evaluation.
\\
\\
As shown in the figure, the training and validation loss for configuration 2 decrease gradually and in a similar manner, without any sudden drops, indicating the absence of slight overfitting.
\begin{figure}[H]
    \centering
    \includegraphics[width=0.8\textwidth]{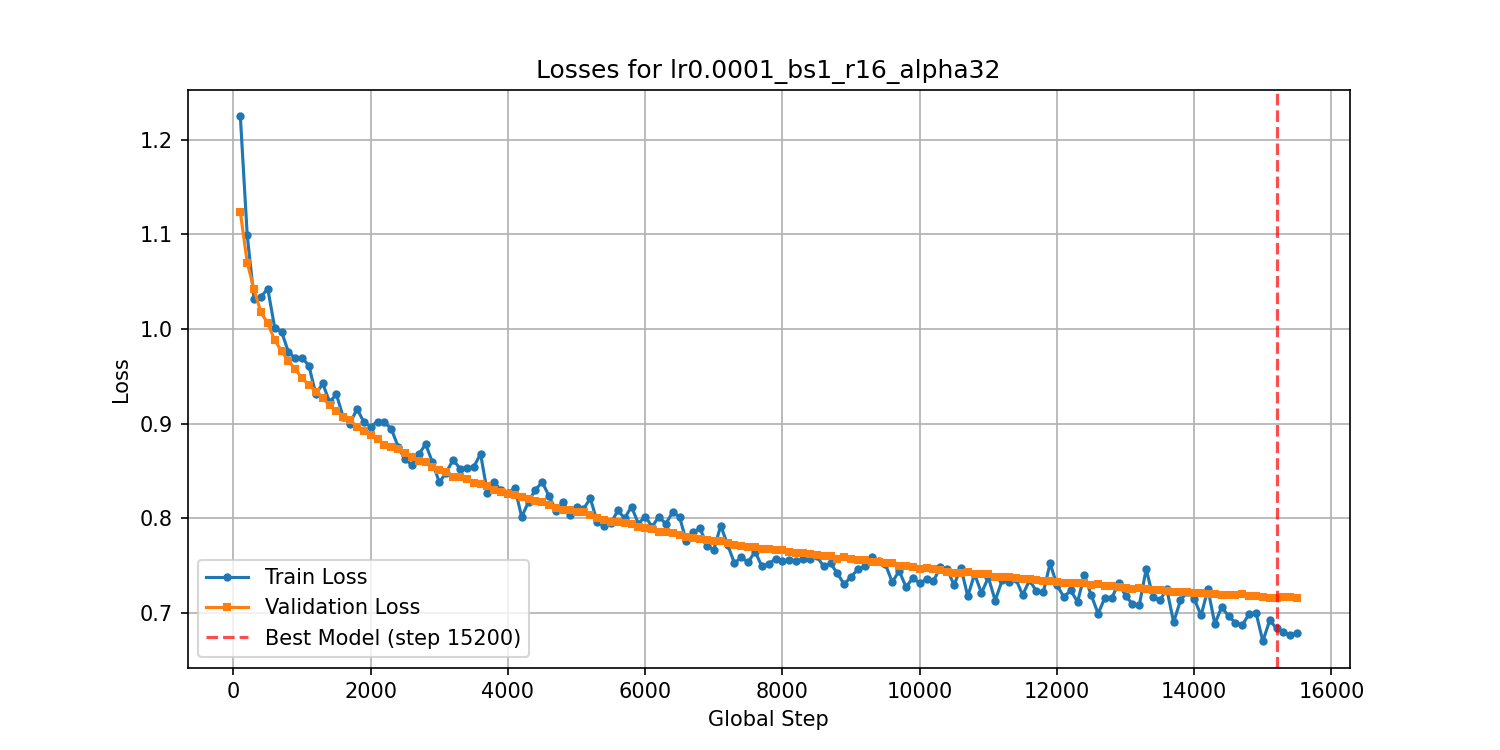}
    \caption{Training and validation loss curves for fine-tuning configuration 2, showing the evolution of the loss over training steps and the convergence behavior of this configuration.}
    \label{fig:lossucurve}
\end{figure}

\textbf{Evaluation distribution shift:} The qualitative prompts used for evaluation may represent a slightly different distribution than the 1,500-example validation set. Fine-tuning configurations with higher loss may better withstand this distribution shift, whereas configurations with very low loss risk overfitting to the validation distribution. This mismatch between distributions could explain the observed discrepancy between quantitative and qualitative evaluation results.
\\

\subsubsection{Implications for Choosing the Best Fine-Tuning Setup}

This divergence highlights critical challenges in LLM evaluation:
\begin{itemize}
    \item \textbf{Validation loss alone does not provide a reliable basis} for selecting among fine-tuning configurations for deployment in conversational AI applications.
    \item \textbf{Human-aligned evaluation} (via LLM-as-a-judge) captures quality dimensions not reflected in perplexity: naturalness, appropriateness, calibration, and pragmatic effectiveness.
    \item \textbf{Multi-metric evaluation frameworks} combining quantitative (loss, perplexity) and qualitative (human/LLM judgment) assessments are essential for real-world deployment.
\end{itemize}

In this context, configuration 2, despite its weaker quantitative performance (loss 0.6807), would be more suitable for deployment scenarios where perceived conversational quality is the primary objective, as it consistently ranks highest in qualitative evaluations by both independent judges. Conversely, for use cases prioritizing computational efficiency, configuration 8 represents a viable alternative, accepting a moderate qualitative trade-off in exchange for lower energy consumption and a lower validation loss.

\subsection{Energy Consumption Analysis}

\begin{figure}[H]
\centering
\includegraphics[width=0.9\linewidth]{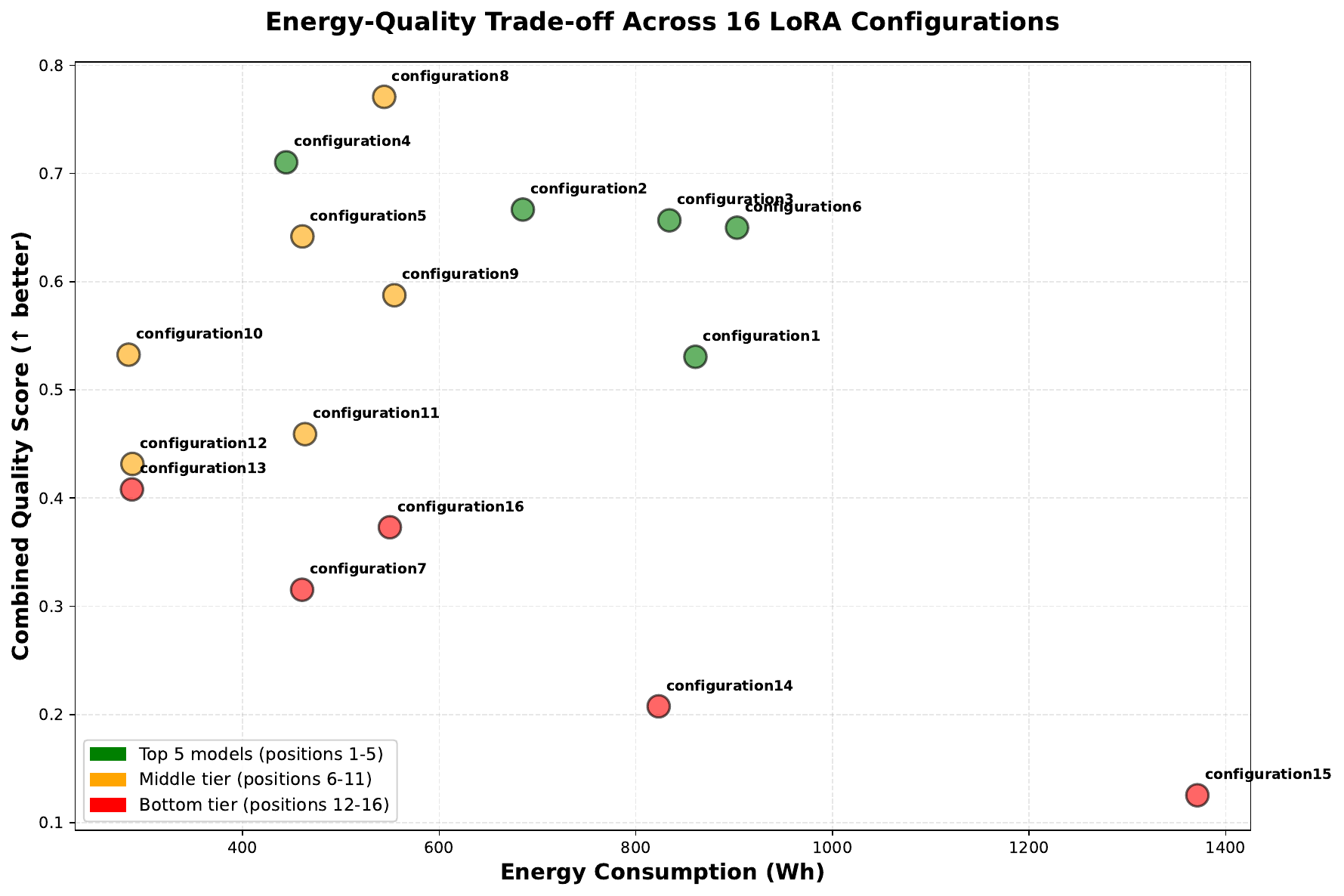}
\caption{Energy-quality trade-off across 16 LoRA configurations.\\[0.3cm]
Each point corresponds to a different fine-tuning configuration of the model. The y-axis corresponds to the combined quality score, calculated from the normalized rankings of GPT-5.2 and Claude 4.5 Sonnet, plus the normalized loss, all divided by 3. The x-axis shows the energy consumed during training. The color of each point indicates the configuration’s ranking in the qualitative evaluation (green = top 5, orange = mid-range, red = bottom 5).
\\
\\
It can be seen that the training of the fourth fine-tuning configuration represents a good compromise between quality and energy consumption.}
\label{fig:energy_quality}
\end{figure}

\begin{table}[H]
\centering
\caption{Complete LoRA configurations and energy consumption during training. Duration format: hours:minutes:seconds.}
\label{tab:complete_results_full}
\resizebox{\textwidth}{!}{%
\begin{tabular}{@{}cccccccccc@{}}
\toprule
\textbf{Configuration} & \textbf{Batch} & \textbf{Grad. Acc.} & \textbf{LR} & \textbf{r} & \textbf{Alpha} & \textbf{Modules} & \textbf{Duration} & \textbf{Avg Power (W)} & \textbf{Energy (Wh)} \\
\midrule
configuration 1  & 1 & 8  & 1e-4 & 16 & 32 & 2 (q,v)       & 2h 45m 59s & 311.08 & 860.59 \\
configuration 2  & 1 & 4  & 1e-4 & 16 & 32 & 2 (q,v)       & 2h 05m 49s & 326.68 & 685.07 \\
configuration 3  & 2 & 4  & 1e-4 & 16 & 32 & 2 (q,v)       & 2h 58m 00s & 281.78 & 834.12 \\
configuration 4  & 2 & 4  & 1e-4 & 16 & 32 & 4 (q,k,v,o)   & 1h 34m 37s & 281.86 & 444.34 \\
configuration 5  & 1 & 8  & 2e-4 & 16 & 32 & 4 (q,k,v,o)   & 1h 38m 19s & 281.82 & 460.88 \\
configuration 6  & 1 & 8  & 5e-5 & 16 & 32 & 4 (q,k,v,o)   & 3h 12m 28s & 281.77 & 902.92 \\
configuration 7  & 1 & 8  & 1e-4 & 16 & 32 & 4 (q,k,v,o)   & 1h 38m 21s & 281.82 & 460.52 \\
configuration 8  & 1 & 8  & 1e-4 & 16 & 32 & 7 (all)       & 1h 55m 50s & 281.80 & 544.04 \\
configuration 9  & 1 & 8  & 5e-5 & 16 & 32 & 7 (all)       & 1h 58m 03s & 281.78 & 554.40 \\
configuration 10 & 1 & 8  & 2e-4 & 16 & 32 & 7 (all)       & 1h 00m 36s & 281.59 & 284.08 \\
configuration 11 & 1 & 8  & 1e-4 & 32 & 64 & 4 (q,k,v,o)   & 1h 38m 41s & 281.85 & 463.52 \\
configuration 12 & 1 & 8  & 1e-4 & 32 & 64 & 7 (all)       & 1h 01m 34s & 281.55 & 287.97 \\
configuration 13 & 1 & 8  & 2e-4 & 32 & 64 & 7 (all)       & 1h 01m 36s & 281.62 & 287.49 \\
configuration 14 & 1 & 16 & 1e-4 & 16 & 32 & 2 (q,v)       & 2h 55m 19s & 281.81 & 823.18 \\
configuration 15 & 1 & 32 & 1e-4 & 16 & 32 & 2 (q,v)       & 4h 52m 37s & 281.74 & 1371.13 \\
configuration 16 & 1 & 8  & 5e-5 & 32 & 64 & 7 (all)       & 1h 56m 58s & 282.01 & 549.81 \\
\bottomrule
\end{tabular}%
}
\end{table}

\subsubsection{Key Energy-Performance Observations}

\textbf{Energy varies 5$\times$ across configurations:} configuration 10 consumes 284 Wh (most efficient) while configuration 15 consumes 1371 Wh (least efficient)—a 4.8$\times$ difference. This demonstrates that hyperparameter choices impact training energy.
\\
\\
\textbf{More modules do not necessarily consume more energy:} Comparing, configuration 1 (2 modules, 861 Wh) versus configuration 8 (7 modules, 544 Wh), the configuration targeting more modules consumes less energy despite higher per-step computational cost. This paradox is explained by early stopping: configuration 8 converges faster (1h 56m) than configuration 1 (2h 46m), terminating earlier and thus consuming less total energy. Fast convergence with appropriate hyperparameters dominates per-step costs.
\\
\\
\textbf{Quality-energy Pareto frontier:} Figure~\ref{fig:energy_quality} reveals several Pareto-optimal configurations: configuration 2 (best qualitative, 685 Wh), configuration 8 (best combined quality score, 544 Wh), and configuration 4 (3rd qualitative, 444 Wh). The choice among these depends on deployment priorities: pure qualitative performance (configuration 2), balanced quantitative-qualitative with efficiency (configuration 8), or maximum energy efficiency among top qualitative performers (configuration 4).

\section{Conclusion}
\label{sec:conclusion}

We presented a systematic study of parameter-efficient fine-tuning for telecommunications customer support, contributing:
\begin{enumerate}
\item A combinatorial synthetic data generation methodology producing 1,560 distinct problem scenarios,

\item A comprehensive evaluation of 16 LoRA configurations showing that increasing target module coverage is more effective than increasing LoRA rank for improving quantitative performance (loss and perplexity),

\item Evidence that low validation loss alone is insufficient for selecting fine-tuning configurations due to a marked divergence between quantitative and qualitative performance, and

\item An energy-aware evaluation revealing up to a 5× variation in training energy consumption across configurations.
\end{enumerate}
Our main finding is the observed divergence between quantitative and qualitative performance: the fine-tuning configuration with the highest validation loss (configuration 2, 0.6807) achieves the best qualitative evaluations according to GPT-5.2 and Claude 4.5 Sonnet, while the one with the lowest loss (configuration 8, 0.5024) ranks only 6th–7th qualitatively. These results indicate that qualitative evaluation is essential for deploying conversational systems, as validation loss alone does not fully capture the perceived quality of responses.

From a strictly quantitative perspective, increasing the number of LoRA target modules (Q, K, V, O, gate, up, down) is more beneficial than increasing LoRA rank. Extending adaptation to all seven modules produces greater improvements in loss than increasing the rank from 16 to 32. However, this additional parameter capacity would likely be better utilized with a more complex dataset than the one used in this study—particularly with longer and more diverse interactions—so that the added parameters support richer generalization rather than overfitting relatively simple patterns.

For practitioners, our results suggest:
\begin{enumerate}
\item Using r = 16 for conversational fine-tuning of a 3B-parameter model,
\begin{enumerate}
\item targeting only the attention modules (Q, V, K, O) when the dataset is relatively simple or limited in diversity,

\item and extending adaptation to the feed-forward layers (gate, up, down) when the dataset is more complex, longer, or more varied, to leverage the additional adaptation capacity;
\end{enumerate}
\item Incorporating qualitative evaluation (human or LLM-as-a-judge) alongside quantitative metrics in fine-tuning configuration selection pipelines;

\item Including energy measurements to identify efficiency–performance trade-offs.

\end{enumerate}

\subsection{Future Work}
Several directions can extend and strengthen the present study.
\\
\\
\textbf{Human evaluation:} First, a human-centered qualitative evaluation should be conducted. While LLM-as-a-judge provides a scalable and cost-effective proxy, involving domain experts (e.g., telecommunications support specialists) as well as real customers would offer a more robust assessment of response quality. Expert evaluation could focus on technical correctness, procedural safety, and regulatory compliance, whereas customer evaluation could emphasize clarity, usefulness, tone, and perceived helpfulness. 
\\
\\
\textbf{Complex and diverse dataset:} Second, future work should explore the generation of a more complex and diverse dataset. The current synthetic dataset, although systematically constructed, remains limited in linguistic variability. Expanding it to include longer multi-turn interactions, richer domain-specific terminology, edge cases, ambiguous problem descriptions, and varied user profiles would better reflect real customer support scenarios. A more complex dataset would also allow a more meaningful evaluation of higher-capacity LoRA configurations, potentially leveraging feed-forward layer adaptation to enhance generalization rather than overfitting to relatively simple patterns.
\\
\\
Together, these directions would contribute to a more comprehensive evaluation framework, bridging quantitative optimization, qualitative assessment, and real-world deployment constraints.
\bigskip
\bigskip
\bigskip
\bigskip
\bigskip
\bigskip
\bigskip
\bigskip
\bigskip
\bigskip
\bigskip
\bigskip
\bigskip
\bigskip
\bigskip
\bigskip
\bigskip
\bigskip
\bigskip
\bigskip
\bigskip
\bigskip
\bigskip
\bigskip
\bigskip
\bigskip
\bigskip
\bigskip
\bigskip
\bigskip
\bigskip
\bigskip

\bibliographystyle{plainnat}
\bibliography{references}

@article{yang2024qwen25,
  title={Qwen2.5: A party of foundation models},
  author={Yang, An and Yang, Baosong and Hui, Binyuan and Zheng, Bo and Yu, Bowen and Zhou, Chang and Li, Chengpeng and Li, Chengyuan and Liu, Dayiheng and Huang, Fei and others},
  journal={arXiv preprint arXiv:2412.15115},
  year={2024}
}

@article{hu2021lora,
  title={LoRA: Low-rank adaptation of large language models},
  author={Hu, Edward J and Shen, Yelong and Wallis, Phillip and Allen-Zhu, Zeyuan and Li, Yuanzhi and Wang, Shean and Wang, Lu and Chen, Weizhu},
  journal={arXiv preprint arXiv:2106.09685},
  year={2021}
}

@article{zhang2023adalora,
  title={Adaptive budget allocation for parameter-efficient fine-tuning},
  author={Zhang, Qingru and Chen, Minshuo and Bukharin, Alexander and He, Pengcheng and Cheng, Yu and Chen, Weizhu and Zhao, Tuo},
  journal={arXiv preprint arXiv:2303.10512},
  year={2023}
}

@article{dettmers2023qlora,
  title={QLoRA: Efficient finetuning of quantized LLMs},
  author={Dettmers, Tim and Pagnoni, Artidoro and Holtzman, Ari and Zettlemoyer, Luke},
  journal={arXiv preprint arXiv:2305.14314},
  year={2023}
}

@article{li2021prefix,
  title={Prefix-tuning: Optimizing continuous prompts for generation},
  author={Li, Xiang Lisa and Liang, Percy},
  journal={arXiv preprint arXiv:2101.00190},
  year={2021}
}

@article{lester2021prompt,
  title={The power of scale for parameter-efficient prompt tuning},
  author={Lester, Brian and Al-Rfou, Rami and Constant, Noah},
  journal={arXiv preprint arXiv:2104.08691},
  year={2021}
}

@article{wang2023selfinstruct,
  title={Self-instruct: Aligning language models with self-generated instructions},
  author={Wang, Yizhong and Kordi, Yeganeh and Mishra, Swaroop and Liu, Alisa and Smith, Noah A and Khashabi, Daniel and Hajishirzi, Hannaneh},
  journal={arXiv preprint arXiv:2212.10560},
  year={2023}
}

@article{taori2023alpaca,
  title={Alpaca: A strong, replicable instruction-following model},
  author={Taori, Rohan and Gulrajani, Ishaan and Zhang, Tianyi and Dubois, Yann and Li, Xuechen and Guestrin, Carlos and Liang, Percy and Hashimoto, Tatsunori B},
  journal={Stanford Center for Research on Foundation Models},
  year={2023}
}

@article{strubell2019energy,
  title={Energy and policy considerations for deep learning in NLP},
  author={Strubell, Emma and Ganesh, Ananya and McCallum, Andrew},
  journal={arXiv preprint arXiv:1906.02243},
  year={2019}
}

@article{patterson2021carbon,
  title={Carbon emissions and large neural network training},
  author={Patterson, David and Gonzalez, Joseph and Le, Quoc and Liang, Chen and Munguia, Lluis-Miquel and Rothchild, Daniel and So, David and Texier, Maud and Dean, Jeff},
  journal={arXiv preprint arXiv:2104.10350},
  year={2021}
}

@article{schwartz2020green,
  title={Green AI},
  author={Schwartz, Roy and Dodge, Jesse and Smith, Noah A and Etzioni, Oren},
  journal={Communications of the ACM},
  volume={63},
  number={12},
  pages={54--63},
  year={2020}
}

@article{geva2021transformer,
  title={Transformer feed-forward layers are key-value memories},
  author={Geva, Mor and Schuster, Roei and Berant, Jonathan and Levy, Omer},
  journal={arXiv preprint arXiv:2012.14913},
  year={2021}
}

@article{loshchilov2019adamw,
  title={Decoupled weight decay regularization},
  author={Loshchilov, Ilya and Hutter, Frank},
  journal={arXiv preprint arXiv:1711.05101},
  year={2019}
}

@article{wolf2020transformers,
  title={Transformers: State-of-the-art natural language processing},
  author={Wolf, Thomas and Debut, Lysandre and Sanh, Victor and Chaumond, Julien and Delangue, Clement and Moi, Anthony and Cistac, Pierric and Rault, Tim and Louf, R{\'e}mi and Funtowicz, Morgan and others},
  journal={Proceedings of the 2020 Conference on Empirical Methods in Natural Language Processing: System Demonstrations},
  pages={38--45},
  year={2020}
}

@misc{mangrulkar2022peft,
  title={PEFT: State-of-the-art parameter-efficient fine-tuning methods},
  author={Mangrulkar, Sourab and Gugger, Sylvain and Debut, Lysandre and Belkada, Younes and Paul, Sayak},
  year={2022},
  howpublished={\url{https://github.com/huggingface/peft}}
}

@misc{vonwerra2022trl,
  title={TRL: Transformer reinforcement learning},
  author={von Werra, Leandro and Belkada, Younes and Tunstall, Lewis and Beeching, Edward and Thrush, Tristan and Lambert, Nathan},
  year={2022},
  howpublished={\url{https://github.com/huggingface/trl}}
}

@misc{google2024gemini,
  title={Gemini: A family of highly capable multimodal models},
  author={Google DeepMind},
  year={2024},
  note={Technical Report}
}

@article{baysan2025llmjudge,
  title={{LLM-as-a-Judge}: automated evaluation of search query parsing using large language models},
  author={Baysan, Mehmet Selman and Uysal, Serkan and {\.I}{\c{s}}lek, {\.I}rem and {\c{C}}{\i\u{g}} Karaman, {\c{C}}a\u{g}la and G{\"u}ng{\"o}r, Tunga},
  journal={Frontiers in Big Data},
  year={2025}
}

@article{zheng2023judging,
  title={Judging {LLM}-as-a-{J}udge with {MT}-{B}ench and {C}hatbot {A}rena},
  author={Zheng, Lianmin and Chiang, Wei-Lin and Sheng, Ying and Zhuang, Siyuan and Wu, Zhanghao and Zhuang, Yonghao and Lin, Zi and Li, Zhuohan and Li, Dacheng and Xing, Eric P and Zhang, Hao and Gonzalez, Joseph E and Stoica, Ion},
  journal={arXiv preprint arXiv:2306.05685},
  year={2023}
}

\end{document}